\documentclass[conference]{IEEEtran}
\IEEEoverridecommandlockouts
\usepackage{cite}
\usepackage{amsmath,amssymb,amsfonts}
\usepackage{algorithmic}
\usepackage{graphicx}
\usepackage{textcomp}
\usepackage{xcolor}
\usepackage{hyperref}
\usepackage{caption}
\usepackage{subcaption}
\usepackage{tabularx}
\usepackage{booktabs}
\usepackage[a4paper, total={184mm,239mm}]{geometry}
\def\BibTeX{{\rm B\kern-.05em{\sc i\kern-.025em b}\kern-.08em
    T\kern-.1667em\lower.7ex\hbox{E}\kern-.125emX}}
\begin{document}

\title{Robustifying the Deployment of tinyML Models for Autonomous mini-vehicles}

 \author{\IEEEauthorblockN{Miguel de Prado
 \IEEEauthorrefmark{1}\IEEEauthorrefmark{3}}
 \IEEEauthorblockA{miguel.deprado@he-arch.ch}
 \and
  \IEEEauthorblockN{Manuele Rusci
 \IEEEauthorrefmark{2}}
 \IEEEauthorblockA{manuele.rusci@unibo.it}
 \and
 \IEEEauthorblockN{Romain Donze
 \IEEEauthorrefmark{1}
 }
 \IEEEauthorblockA{romain.donze@he-arc.ch}
 \and
 \IEEEauthorblockN{Alessandro Capotondi
 \IEEEauthorrefmark{4}}
 \IEEEauthorblockA{alessandro.capotondi@unimore.it}
 \and
 \IEEEauthorblockN{\hspace{2cm}Serge Monnerat
 \IEEEauthorrefmark{1}}
 \IEEEauthorblockA{\hspace{2cm}Serge.Monnerat@he-arc.ch}
 \and
 \IEEEauthorblockN{\hspace{-4cm}Luca Benini
 \IEEEauthorrefmark{2}\IEEEauthorrefmark{3}}
 \IEEEauthorblockA{\hspace{-4cm}lbenini@iis.ee.ethz.ch}
 \and
 \IEEEauthorblockN{\hspace{-5cm}Nuria Pazos
 \IEEEauthorrefmark{1}}
 \IEEEauthorblockA{\hspace{-5cm}Nuria.Pazos@he-arc.ch}
 \and
 \IEEEauthorblockA{\hspace{1cm}\IEEEauthorrefmark{1} He-Arc Ingenierie, HES-SO
 \IEEEauthorrefmark{2}DEI, University of Bologna
 \IEEEauthorrefmark{3}Integrated System Lab, ETH Zurich
 \IEEEauthorrefmark{4}UNIMORE}
}
\maketitle

\begin{abstract}
Standard-size autonomous navigation vehicles have rapidly improved thanks to the breakthroughs of deep learning. However, scaling autonomous driving to low-power systems deployed on dynamic environments poses several challenges that prevent their adoption. 
To address them, we propose a closed-loop learning flow for autonomous driving mini-vehicles 
that includes the target environment in-the-loop. 
We leverage a family of compact and high-throughput tinyCNNs to control the mini-vehicle, which learn in the target environment by imitating a computer vision algorithm, i.e., the expert. Thus, the tinyCNNs, having only access to an on-board fast-rate linear camera, gain robustness to lighting conditions and improve over time.
Further, we leverage GAP8, a parallel ultra-low-power RISC-V SoC, to meet the inference requirements. When running the family of CNNs, our GAP8's solution outperforms any other implementation on the STM32L4 and NXP k64f (Cortex-M4), reducing the latency by over 13x and the energy consummation by 92\%.
\end{abstract}


\vspace{-0.2cm}
\section{Introduction}
Autonomous driving has made giant strides since the advent of deep learning (DL). However, scaling this technology to micro- and nano- vehicles poses severe challenges in terms of functionality and robustness due to their limited computational and memory resources of the on-board processing unit~\cite{palossi201964mw, flops}. 

Micro-Controller Units (MCUs) are typically chosen on small unmanned vehicles to balance the mW power cost of the sensing front-end and keep the system-energy low to extend the battery life. Traditionally, driving decisions have been off-loaded and carried out remotely, implying energy-expensive, long-latency, and unreliable transmissions of raw data to cloud servers~\cite{data_move}. Off-system transfers can be prevented by processing data on-board and directly driving the motor controllers. 
Thus, tiny machine learning (TinyML) has appeared as a new field to address these challenges and tackle on-device sensor data analysis at hardware, algorithmic and, software level \cite{tinyML,banbury2020benchmarking}.

A major challenge for the design of autonomous driving decision models is that the real-world environment change over time: the data distribution that has been initially learned might not match the underlying distribution of the current environment, e.g., the car driving through different landscapes or weather conditions. Hence, there is an increasing need to adapt to ever-changing environments to make vehicles more robust and efficient over time.

\subsection{Goal specification: Robust low-power Autonomous Driving} \label{problem}

\begin{figure}[!t]
  \centering
  \includegraphics[width=3.5in, height=1.3in]{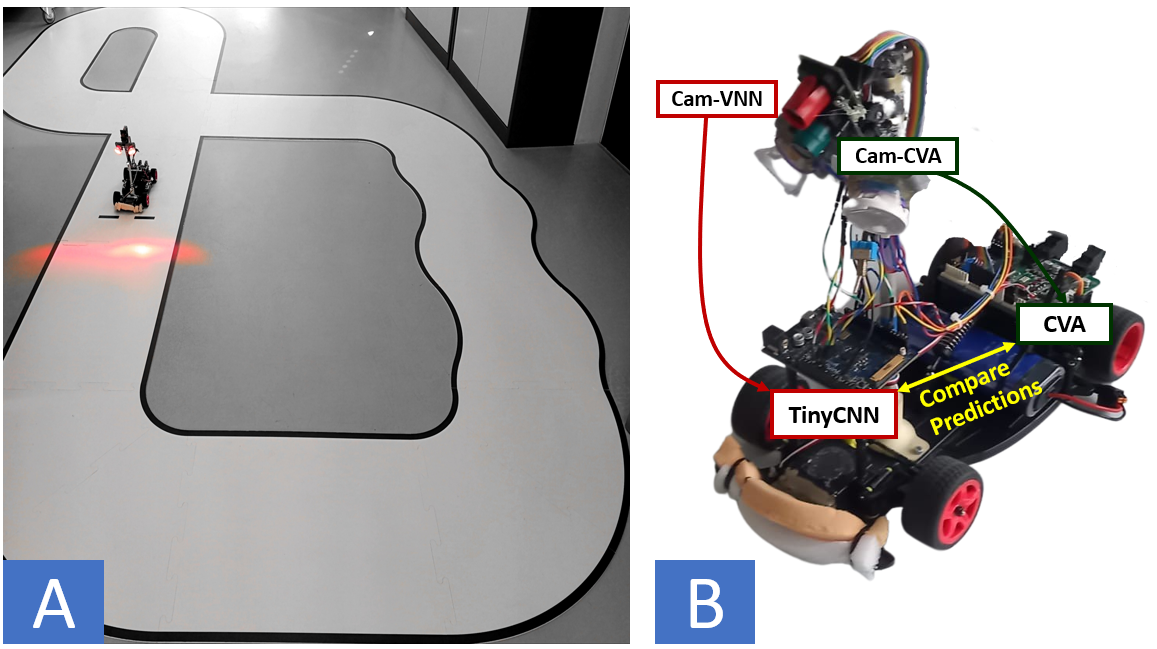}
  \caption{\small\textbf{Automotive application use-case.} A) Mini-vehicle running on circuit track. B) Close-loop learning task. One linear camera feeds the GAP8 (tinyCNN) and another the NXP k64f (CVA).}
  \label{fig:circuit}
  \vspace{-0.4cm}
\end{figure}


We aim to shed light in the robustification of DL methods for autonomous systems deployed on dynamic environments. We specifically focus on enabling the deployment of tinyCNNs to a low-power autonomous driving vehicle. 
We base our design on the NXP cup framework \cite{NXPcup}, an autonomous racing competition that offers a solid ground to test new ideas that can be reproduced, ported to other autonomous devices, e.g., drones, or scale up to bigger vehicles. 
Our vehicle consists of a battery-powered mini-car that needs to detect autonomously 7 different states in a controlled circuit: 
\textit{GoStraight, TurnLeft, TurnRight, CrossingStreets, StartSpeedLimit and StopSpeedLimit} as shown in Fig.~\ref{fig:circuit}.
The vehicle contains a linear camera and an on-board NXP k64f  MCU~\cite{NXPK64F} to detect the current state, compute the required action and drive the actuators accordingly. 

The process was originally based on a conventional and handcrafted computer-vision algorithm (named as CVA) that predicts accurately only under stable light conditions. The fragility to light condition is due to the nature of the CVA as it calculates derivatives on the input image which requires good contrast. 
The lack of robustness to light conditions is countered by setting the camera with a variable acquisition time that adapts to the lighting conditions (controlled by a PID) to obtain clear images. However, the variability and length of the time ($\geq$2 ms) limit the vehicle's agility.
Thus, we take a tinyML approach and aim at replacing the conventional CVA by a tinyCNN to: \textit{i)} improve the robustness to lighting variations, and \textit{ii)} increase the performance, i.e., actions/sec, by learning challenging features from short and constant acquisition times.

\subsection{Contributions}
We present an end-to-end flow of data, algorithms, and deployment tools that facilitates the deployment and enhance the robustness of a family of tinyCNNs for an autonomous low-power mini-vehicle. Thus, our contributions are the following:

\textit{1)} We introduce a closed-loop learning flow that enables the tinyCNNs to learn through demonstration. By imitating an expert with access to good-quality images, the tinyCNNs gain robustness to lighting conditions while having only access to a fast-rate camera, thereby doubling the system's throughput.

\vspace{0.05cm}

\textit{2)} We leverage GAP8~\cite{flamand2018gap}, a parallel ultra-low-power RISC-V SoC, to meet the CNN inference computing requirements by adding it as a System-on-Module (SoM) on the NXP platform.

Further, we compare our deployment solution on GAP8 (50 MHz) with two Cortex-M: an STM32L4 (80 MHz) and an NXP k64f (120 MHz). We present the Pareto-optimal front where our solution dominates all other implementations, reducing the latency by over 13x and the energy consumption by 92\%.

    

\section{Related work}\label{sec:related}
We find 3 main topics related to this work in the literature:

\subsubsection{High-Performance Autonomous Driving} 
There exist multiple CNN approaches for autonomous driving~\cite{kuutti2019survey}, ranging from standard-size to small-scale vehicles. On the higher end, Nvidia and Tesla introduced PiloNet~\cite{bojarski2017explaining} and AutoPilot~\cite{AutoPilot}, requiring dedicated platforms such as TESLA FSD chip and NVIDIA drive, which provide TOPs of computing power and tens of GBytes of memory for their large CNN solutions. Other approaches such as DeepRacer~\cite{DeepRacer}, F1/10~\cite{F1/10}, or DonkeyCar~\cite{Donkeycar} require GOPs and hundreds of MBytes that cover using platforms like Nvidia Jetson, Raspberry PI or Intel Atom. By constrast, we focus on end-to-end CNN solutions suitable for very low-power vehicles with MCUs featuring MOPs and up to a few MBytes, which is an unexplored field.

\subsubsection{Learning Methodology} 
Imitation Learning (IL) leverages the idea of a student learning from an expert that gives directives through demonstration~\cite{imitation_learning,imitation_learning2}. 
In this context, ALVINN \cite{pomerleau1989alvinn} proposed a CNN-based system that 
learns to infer the steering angle from images taken from a camera on-board. Similarly, PilotNet~\cite{bojarski2017explaining} and J-Net~\cite{kocic2019end} employed a system that collects the driver's signal to label a training dataset with on-board cameras. However, these approaches only use the expert to label the training datasets. Instead, we propose a closed-loop learning flow where the learner confronts the expert in real drive and gradually improves through demonstration. In this direction, Pan et al. proposed \cite{pan2018agile} where they optimize the policy (online) of an reinforcement learning agent that imitates an expert driver with access to costly resources, while the agent has only access to economic sensors. However, their approach is not compatible with our use case where low-power systems cannot hold such level of computation and memory. 

\subsubsection{Low-power DL deployment}
Multiple software stacks have been introduced on inference tasks for resource-constrained MCUs.
In this context, 
STMicroelectronics has released X-CUBE-AI to generate optimized code for low-end STM32 MCUs~\cite{xcubeai}. Similarly, ARM has provided the CMSIS-NN~\cite{lai2018cmsis}, which targets Cortex-M processors and provides a complete backend for quantized DL networks exploiting 2x16-bit vector MACs instructions~\cite{zhang2017hello, chowdhery2019visual}. 
Recently, the CMSIS-NN dataflow has been ported to a parallel low-power architecture, PULP, originating PULP-NN~\cite{garofalo2019pulp}, which exploits 4x8-bit SIMD MAC instructions 
on a parallel processor, such as the GAP8. 
In this work, we leverage the GAP8 for the autonomous driving use case and provide a quantitative energy-latency-quality comparison of these solutions. 

\section{Robust navigation with tinyML} \label{sc:robust}
We aim to take a tinyML approach and replace the initial CVA solution by a tinyCNN. First, we evaluate an initial setup and verify the challenge of porting DL methods on MCUs. 

\subsection{Initial Evaluation and Challenges}
\label{sec:CL}
\subsubsection{Data collection}
We manually collect and label 3 initial datasets, each containing around 1000 samples per class (driving action) for the training set and 300 for the test set. The first dataset, \textit{Dset-2.0}, contains samples with a fixed acquisition time of 2.0~ms - clear enough images - where the CVA can still predict well the required action. On the other hand, the second and third datasets, \textit{Dset-1.5} and \textit{Dset-1.0}, hold more challenging samples (low contrast) with 1.5~ms and 1.0~ms acquisition times where the CVA fails to predict well, and thus, we aim to use a CNN instead.

\subsubsection{Training} \label{In:Training}
We choose LeNet5~\cite{lecun2015lenet} for our initial evaluation as it is small and well-known CNN architecture, which is also used in~\cite{kocic2019end}.
We use PyTorch as training environment with cross-entropy loss function, SGD optimizer, data augmentation and dropout to avoid over-fitting. Thus, we obtain an accuracy of 99.53\% on the \textit{Dset-2.0} test set, but only 84.12\% and 81.27\% on the more challenging \textit{Dset-1.5} and \textit{Dset-1.0} test sets. 


%

\newcolumntype{b}{X}
\newcolumntype{s}{>{\hsize=.35\hsize}X}

\begin{table}[t]
    \centering
    \begin{tabularx}{\columnwidth}{bsssss}
        & \textbf{LeNet5} & \textbf{VNN4} & \textbf{VNN3} & \textbf{VNN2} & \textbf{VNN1}\\
        \toprule
        \#~Parameters (K) & 72,85  & 6.04  & 0.97 & 1.29 & 0.48 \\
        Complexity(KMAC) & 181.25 & 163.41 & 28.69 & 5.82 & 7.5 \\
    \end{tabularx}
    \caption{\small\textbf{Vehicle Neural Network (VNN) family}}
    \label{network_table}
    \vspace{-0.5cm}
\end{table}

\subsubsection{Deployment}
We employ CMSIS-NN (Int8) as a backend to execute LeNet5 on the NXP k64f. The execution time turns out to be 5.4~ms, far too long compared to the $\approx$2~ms achieved by the conventional CVA on the same platform and conditions.

\vspace{0.1cm}
Given LeNet5's fragility to lighting conditions and its long latency, the initial CNN setup provides no benefit compared to the original CVA. To address these challenges, first, we create a family of tinyCNNs for efficient MCU deployment. Next, we introduce the GAP8 as a SoM to accelerate the CNN inference and finally, we detail the closed-loop learning methodology.

\subsection{Vehicle Neural Network (VNN) Family}
We modify LeNet5's topology by varying the number of convolutional (conv) layers and the stride to shrink the model size, the number of operations, and the latency. To have a higher tolerance to the diffusion of features in low-light conditions, we opt for a relatively large kernel size (k=5) for both the conv. and the pooling layers, having the latter a stride of 3 to reduce the number of activations.
As a result, we have created a family of tinyCNN called Vehicle Neural Networks (VNNs), see Table~\ref{network_table}, containing a range of layers that span from 1 to 3 convs. followed by one fully-connected layer for the final classification.
Networks and datasets have been open-sourced\footnote{ \textit{https://github.com/praesc/Robust-navigation-with-TinyML}}

\subsection{System-on-Module (SoM) setup}
We leverage GAP8~\cite{flamand2018gap}, a parallel ultra-low-power RISC-V SoC, to meet the CNN's inference computing requirements (we describe GAP8 and efficient deployment in Section~\ref{sec:CL:deployment}). We add GAP8 as a SoM on the NXP platform and set up the system with two synchronized cameras (calibrated to have the same view of the circuit), one feeding the NXP and another feeding the GAP8, see Fig. \ref{fig:circuit}. We use the  GAP8 as a CNN accelerator, which is only in charge of inferring the VNN on the input image, while the NXP controls all the other sensors and actuators. The results from the VNN are constantly transferred via UART from the GAP8 to the NXP to control the actuators.

\subsection{Closed-loop Learning System} \label{sec:CL:IL}

Initially, our family of VNNs achieves an accuracy of 78\%-83\% on \textit{Dset-1.0}, a drop of 15\%-20\% compared to the VNNs trained on the \textit{Dset-2.0} setup. Thus, we need a learning strategy to enhance the VNNs' robustness on the low-contrast images from the \textit{Dset-1.0} setup. Hence, we propose a closed-loop learning system as a way to gradually improve the quality of our datasets and boost the VNNs' accuracy. 

\begin{figure}[t]
  \centering
  \includegraphics[width=3.5in, height=1.0in]{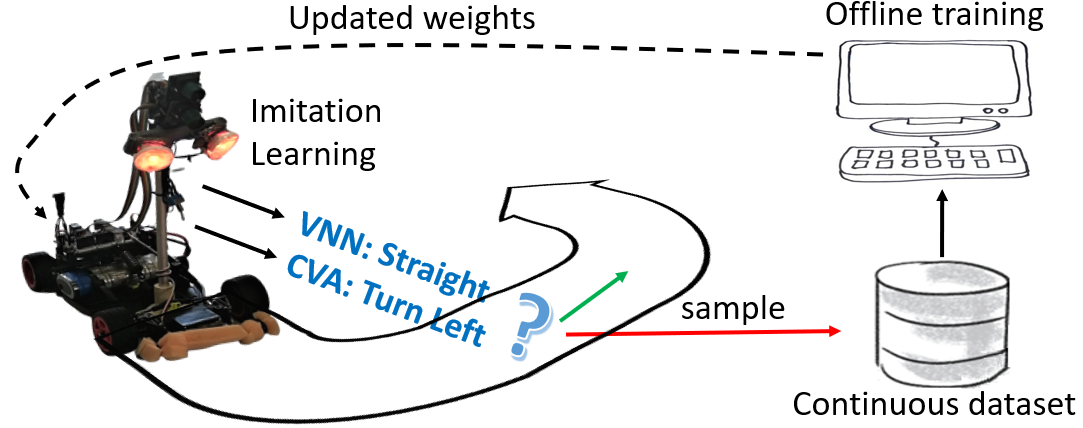}
  \caption{\small\textbf{Closed-loop learning pipeline.} End-to-end closed-loop learning cyclic methodology via Imitation Learning.}
  \label{fig:pipeline}
  \vspace{-0.5cm}
\end{figure}

We implement this technique by collecting valuable data from the sensors at runtime, training (offline) the model from scratch on the cumulative set of data, and pushing back the updates to the deployed VNN, see Fig.~\ref{fig:pipeline}. Thus, we pursue two main objectives: \textit{i)} improve the robustness to lighting variations, and \textit{ii)} increase the performance, i.e., actions/sec, by learning challenging features from shorter acquisition times.

We have experimentally tested\footnote{Not possible to benchmark CVA statically on our \textit{Dsets} as it uses previous samples to predict the current one, i.e., it works on a continuous data stream.} that the CVA accurately predicts provided adequate light conditions. We can assume the predictions of the CVA as ground-truth and follow an imitation learning (IL) approach where we use the CVA as a teacher to help the VNNs learn better features.
Hence, we decouple the original system and propose a setup with two cameras:
\begin{itemize}
    \item Cam-CVA: set with a variable (and long) acquisition time that adapts to the lighting conditions (controlled by a PID) and always provides clear images. This camera feeds the CVA running on the NXP.
    \item Cam-VNN: set with a short and constant acquisition time that captures the lighting variability of the environment. This camera feeds the VNN on the GAP8.
\end{itemize}
By confronting the results of both algorithms while the mini-vehicle runs in a lighting-changing environment, we can improve the generalization capacity of the deployed VNN. Moreover, we can also leverage IL to increase the VNNs' accuracy on the more challenging \textit{Dset-1.5} and \textit{Dset-1.0} setups, which, in turn, improves the system's throughput (actions/sec). We carry out the IL methodology at runtime and collect those input images that lead VNN's predictions to differ from those of the CVA. The samples get automatically labeled by the CVA and are sent over to the PC. We include the new samples in the training set to reinforce those classes where the VNN has failed. 
The learning procedure can be repeated multiple times. However, we find that the frequency of new sample discovery decreases over time as the deployed solution progressively improves. Thus, we show an experiment with several phases.

\subsection{Experimental setup and Results} \label{sec:results}
We demonstrate the closed-loop learning methodology by first training a VNN on \textit{Dset-1.0}, the most challenging setup.
Next, we combine our three \textit{Dsets: 2.0, 1.5, and 1.0}, and train the VNNs on it to make them have access to richer data distribution. Then, we deploy a VNN on the mini-vehicle and set \textit{Cam-VNN}'s acquisition time to \textit{1 ms}. We make the vehicle run in a varying-light environment while the VNN and CVA results are confronted. Those samples where the VNN failed are collected at every stage and merged to \textit{Dset-1.0} training set to conform an incremental dataset that we show in 3 phases: \textit{+25\%}, \textit{+50\%} and \textit{+100\%} (new samples wrt. the original \textit{Dset-1.0}). 
In the last stage, i.e., \textit{+100\%}, we heavily alter the light conditions of the environment to improve the robustness of the VNN. We train 
the VNN from scratch at each phase for 1000 epochs on the complete set and send an update of the weights to the vehicle before starting the next phase. We perform each training three times to account for the variability in the random initialization.

\begin{figure}[!t]
  \centering
  \includegraphics[width=0.9\linewidth, height=1.5in]{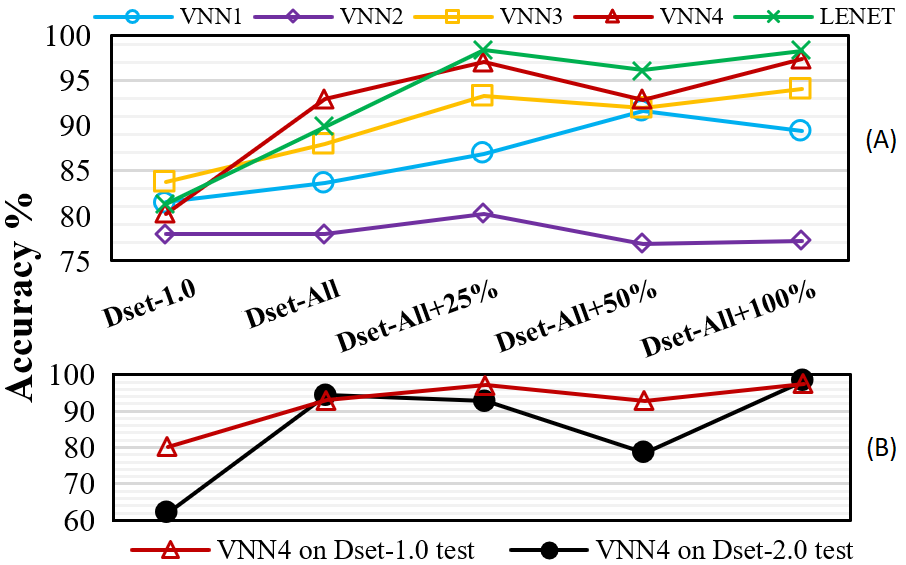}
  \caption{\small \textbf{Closed-loop learning evaluation.} X-axis is shared. A) VNNs accuracy on \textit{Dset-1.0} test (8-bit). B) VNN4 trained on \textit{Dset-1.0} learns/forgets features from \textit{Dset-2.0}.}
  \label{fig:continuous_learning}
  \vspace{-0.5cm}
\end{figure}

Fig.~\ref{fig:continuous_learning}-A displays the results obtained on the \textit{Dset-1.0} test set. Initially, our family of VNNs achieves an accuracy of 78\%-83\%. After training the VNNs on the combined dataset (\textit{Dset-All}), most of the VNNs are able to generalize better, and their accuracy noticeably improves due to the richer diversity of light conditions. Further, when leveraging the closed-loop learning methodology through IL and training the networks on the reinforced dataset, VNN3, VNN4, and LeNet5 reach a top accuracy of 94.1\%, 97.4\%, and 98.3\%.
By contrast, VNN2's accuracy remains mostly constant while VNN1 decays at the end, probably due to their shallow topology and lower capacity, failing to learn from more challenging data. Overall, the closed-loop learning approach allows an increase in accuracy of over 15\% in \textit{Dset-1.0}, matching the accuracy of the CVA on \textit{Dset-2.0} while doubling the throughput of the system.

Fig.~\ref{fig:continuous_learning}-B illustrates how VNN4 learns and forgets features. We train VNN4 on the data from \textit{Dset-1.0} setup, i.e., same as in Fig.~\ref{fig:continuous_learning}-A, and compare VNN4's accuracy on \textit{Dset-1.0} and \textit{Dset-2.0} test sets, which contain samples with different acquisition time and therefore, lighting conditions. We can observe that, at first, when the VNN4 has only been trained on \textit{Dset-1.0} initial's data, the accuracy on \textit{Dset-2.0} is poor. Nonetheless, when VNN4 receives data from \textit{Dset-All}, it generalizes better and performs well in both test sets. However, as the experiment goes on and VNN4 only sees data from the \textit{Dset-1.0} setup, it tends to forget and slowly decreases its accuracy on \textit{Dset-2.0} test set. Finally, by heavily altering the lighting conditions on the \textit{100\%} point, VNN4 can rehearse with data similar to those of \textit{Dset-2.0} and "remembers". This experiment depicts a challenge of continuous learning. It slightly shows the effects of catastrophic forgetting~\cite{culurciello} and how rehearsal techniques can tackle this issue~\cite{lomonaco2019continual}, which we will address in the future.  

\section{MCU efficient deployment} \label{sec:CL:deployment}
We first describe the compression of the VNNs and GAP8's capabilities to achieve an efficient deployment. Then, we offer a comparison between different MCUs when running our VNNs.

\subsection{Network Compression}
Training of CNNs is normally carried out using 
floating-point data types. Such types need specific arithmetic units which may not even be present in MCUs due to their large area and costly energy consumption. Quantization is a compression method that reduces the storage cost of a variable by employing reduced-numerical precision. In addition, low-precision data types can improve the arithmetic intensity of the CNNs by leveraging the instruction-level parallelism. 
Thus, we have quantized our VNN models to fixed-point 8-bit to reduce memory footprint and power consumption~\cite{rusci2018work,rusci2019memory}. We have employed post-training quantization where the weights can be directly quantized while the activations require a validation set (sampled from circuit runs) to determine their dynamic range. Thus, we have observed a low accuracy loss ($<$3\%) on the initial \textit{Dsets} and negligible loss ($<$1\%) after the closed-loop learning phases.

\subsection{MCU Hardware/Software Inference Platforms}
We have leveraged the GAP8~\cite{flamand2018gap}, which is based on the PULP architecture. It includes a RISC-V core, acting as an MCU, and an 8-core RISC-V (cluster) accelerator featuring a DSP-extended ISA that includes SIMD vector instructions such as 4x8-bit Multiply and Accumulate (MAC) operations. Besides, the cluster is equipped with a zero-latency 64kB L1 Tightly Coupled Data Memory and an 512 kB L2 memory.

We compare the GAP8 (at 50 MHz) with two other classes of MCUs: a \textit{low-power single-core} MCU (STM32 L476 at 80  MHz~\cite{stm32l4}), and a \textit{high-performance single-core} MCU (NXP k64f at 120  MHz~\cite{NXPK64F}).
Besides, we also compare different inference backends supported on these devices. We report the GAP8 platform coupled with the PULP-NN~\cite{garofalo2019pulp}, against the STM32 supporting either X-Cube-AI~\cite{xcubeai} or CMSIS-NN~\cite{lai2018cmsis} and, the NXP coupled with the CMSIS-NN~\cite{lai2018cmsis}.

\begin{figure}[t]
  \centering
  \includegraphics[width=\linewidth, height=1.3in]{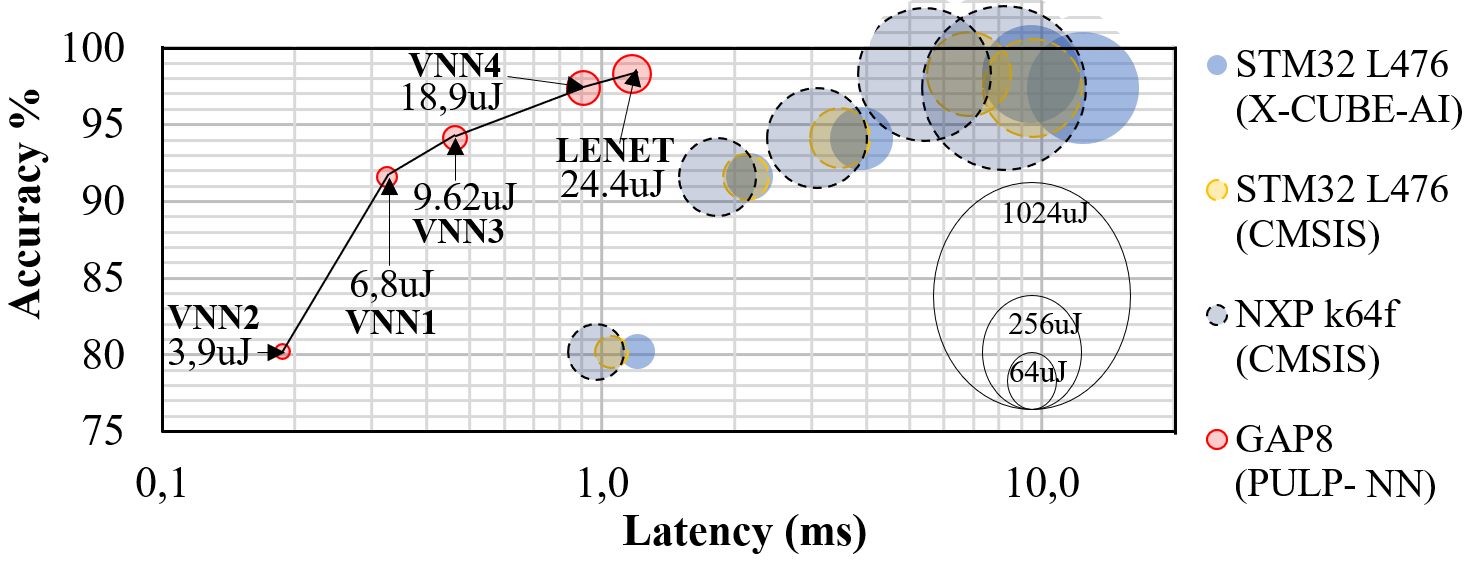}
  \caption{\small \textbf{Accuracy-Latency-Energy Trade-off.} Accuracy (w.r.t. \textit{Dset-1.0}, y-axis), latency (x-axis), and energy-consumption (balloon area) for an inference of the VNNs. 
  Black line highlights Pareto front.}
  \label{fig:accuracy-latency-energy}
  \vspace{-0.5cm}
\end{figure}

\subsection{Energy-Accuracy-Latency Trade-off}
Thanks to the high accuracy obtained through the closed-loop learning methodology, we can now employ the camera with a short-acquisition time ($1~ms$) 
and set it as our latency target for the classification task. 
Fig.~\ref{fig:accuracy-latency-energy} summarizes the accuracy, latency, and energy measured on the different MCUs. All VNNs deployed on GAP8 meet the $1~ms$ target and establish the Accuracy-Latency Pareto frontier, dominating all the other implementations on STM32 L476 and NXP k64f. Remarkably, only VNN2 performs under $1ms$ on the NXP k64f ($0.97 ms$) due to its shallow topology and the higher clock frequency of the NXP. Interestingly, we observe that X-Cube-AI backend is up to 27.8\% slower than CMISIS-NN on the STM32 L476.  

Looking at the energy consumption of a single classification, VNN4 on GAP8 (8 cores, 50 MHz) consumes $18.9\mu J$, -65.2\% less compared to VNN2 on NXP k64f (120MHz) while being over 20\% more accurate. The same VNN2 on GAP8 consumes $3.9\mu J$, -92.8\% less than the NXP k64f. The usage of STM32 L476 (80MHz) is only possible if executing VNN2 and relaxing the target latency up to $1.5 ms$. Yet, it consumes the same amount of energy as VNN4 on GAP8, while the latter provides -13.5\% latency and +17.2\% accuracy.

\section{Conclusion and future work} \label{sec:conclusion}
We have shed light on the robustification of tinyCNN models deployed on a low-power driving use case. We have introduced a closed-loop learning methodology that enables the tinyCNNs to imitate an expert, improving their robustness to lighting conditions in the target deployment scenario. Further, we have proposed a parallel ultra-low-power platform to meet the latency specifications, which consume as little as $3.9\mu J$ for a single inference. we envision that this methodology can be applied for other autonomous use cases, e.g., drones, to improve the robustness and efficiency of tinyML methods. As future work, we aim to perform the training of the CNNs on-chip towards a continuous learning scenario.
\bibliographystyle{IEEEtran}
\bibliography{references}

\begin{thebibliography}{10}
\providecommand{\url}[1]{#1}
\csname url@samestyle\endcsname
\providecommand{\newblock}{\relax}
\providecommand{\bibinfo}[2]{#2}
\providecommand{\BIBentrySTDinterwordspacing}{\spaceskip=0pt\relax}
\providecommand{\BIBentryALTinterwordstretchfactor}{4}
\providecommand{\BIBentryALTinterwordspacing}{\spaceskip=\fontdimen2\font plus
\BIBentryALTinterwordstretchfactor\fontdimen3\font minus
  \fontdimen4\font\relax}
\providecommand{\BIBforeignlanguage}[2]{{%
\expandafter\ifx\csname l@#1\endcsname\relax
\typeout{** WARNING: IEEEtran.bst: No hyphenation pattern has been}%
\typeout{** loaded for the language `#1'. Using the pattern for}%
\typeout{** the default language instead.}%
\else
\language=\csname l@#1\endcsname
\fi
#2}}
\providecommand{\BIBdecl}{\relax}
\BIBdecl

\bibitem{palossi201964mw}
D.~Palossi, A.~Loquercio, F.~Conti, E.~Flamand, D.~Scaramuzza, and L.~Benini,
  ``{A 64mW DNN-based Visual Navigation Engine for Autonomous Nano-Drones},''
  \emph{IEEE Internet of Things Journal}, 2019.

\bibitem{flops}
S.~Bianco, R.~Cadene, L.~Celona, and P.~Napoletano, ``Benchmark analysis of
  representative deep neural network architectures,'' \emph{IEEE Access},
  vol.~6, pp. 64\,270--64\,277, 2018.

\bibitem{data_move}
O.~Mutlu, ``Processing data where it makes sense in modern computing systems:
  Enabling in-memory computation,'' in \emph{2018 7th Mediterranean Conference
  on Embedded Computing (MECO)}.\hskip 1em plus 0.5em minus 0.4em\relax IEEE,
  2018, pp. 8--9.

\bibitem{tinyML}
\BIBentryALTinterwordspacing
``Tinyml,'' 2020. [Online]. Available: \url{https://www.tinyml.org/summit/}
\BIBentrySTDinterwordspacing

\bibitem{banbury2020benchmarking}
C.~R. Banbury, V.~J. Reddi, M.~Lam, W.~Fu, A.~Fazel, J.~Holleman, X.~Huang,
  R.~Hurtado, D.~Kanter, A.~Lokhmotov \emph{et~al.}, ``Benchmarking tinyml
  systems: Challenges and direction,'' \emph{arXiv preprint arXiv:2003.04821},
  2020.

\bibitem{NXPcup}
\BIBentryALTinterwordspacing
``Nxpcup,'' 2020. [Online]. Available: \url{https://nxpcup.nxp.com/}
\BIBentrySTDinterwordspacing

\bibitem{NXPK64F}
\BIBentryALTinterwordspacing
``{NXP K64F},'' 2019. [Online]. Available: \url{https://www.nxp.com}
\BIBentrySTDinterwordspacing

\bibitem{flamand2018gap}
E.~Flamand, D.~Rossi, F.~Conti, I.~Loi, A.~Pullini, F.~Rotenberg, and
  L.~Benini, ``{GAP-8: A RISC-V SoC for AI at the Edge of the IoT},'' in
  \emph{2018 IEEE 29th International Conference on Application-specific
  Systems, Architectures and Processors}.\hskip 1em plus 0.5em minus
  0.4em\relax IEEE, 2018, pp. 1--4.

\bibitem{kuutti2019survey}
S.~Kuutti, R.~Bowden, Y.~Jin, P.~Barber, and S.~Fallah, ``A survey of deep
  learning applications to autonomous vehicle control,'' \emph{arXiv preprint
  arXiv:1912.10773}, 2019.

\bibitem{bojarski2017explaining}
M.~Bojarski, P.~Yeres, A.~Choromanska, K.~Choromanski, B.~Firner, L.~Jackel,
  and U.~Muller, ``Explaining how a deep neural network trained with end-to-end
  learning steers a car,'' \emph{arXiv preprint arXiv:1704.07911}.

\bibitem{AutoPilot}
\BIBentryALTinterwordspacing
``{Auto Pilot},'' 2019. [Online]. Available:
  \url{https://www.tesla.com/autopilot}
\BIBentrySTDinterwordspacing

\bibitem{DeepRacer}
\BIBentryALTinterwordspacing
``{DeepRacer}.'' [Online]. Available: \url{https://aws.amazon.com/deepracer/}
\BIBentrySTDinterwordspacing

\bibitem{F1/10}
M.~O'Kelly, V.~Sukhil, H.~Abbas, J.~Harkins, C.~Kao, Y.~V. Pant, R.~Mangharam,
  D.~Agarwal, M.~Behl, P.~Burgio \emph{et~al.}, ``F1/10: An open-source
  autonomous cyber-physical platform,'' \emph{arXiv preprint arXiv:1901.08567}.

\bibitem{Donkeycar}
\BIBentryALTinterwordspacing
``{DonkeyCar}.'' [Online]. Available: \url{github.com/autorope/donkeycar}
\BIBentrySTDinterwordspacing

\bibitem{imitation_learning}
\BIBentryALTinterwordspacing
``{Introduction to Imitation Learning},'' 2019. [Online]. Available:
  \url{https://blog.statsbot.co/introduction-to-imitation-learning-32334c3b1e7a}
\BIBentrySTDinterwordspacing

\bibitem{imitation_learning2}
\BIBentryALTinterwordspacing
``{ICML 2018: Imitation Learning Tutorial},'' 2018. [Online]. Available:
  \url{https://sites.google.com/view/icml2018-imitation-learning/}
\BIBentrySTDinterwordspacing

\bibitem{pomerleau1989alvinn}
D.~A. Pomerleau, ``Alvinn: An autonomous land vehicle in a neural network,'' in
  \emph{Advances in neural information processing systems}, 1989.

\bibitem{kocic2019end}
J.~Koci{\'c}, N.~Jovi{\v{c}}i{\'c}, and V.~Drndarevi{\'c}, ``An end-to-end deep
  neural network for autonomous driving designed for embedded automotive
  platforms,'' \emph{Sensors}, vol.~19, no.~9, p. 2064, 2019.

\bibitem{pan2018agile}
Y.~Pan, C.-A. Cheng, K.~Saigol, K.~Lee, X.~Yan, E.~Theodorou, and B.~Boots,
  ``Agile autonomous driving using end-to-end deep imitation learning,'' in
  \emph{Robotics: science and systems}, 2018.

\bibitem{xcubeai}
STMicroelectronics, ``{X-CUBE-AI},''
  \url{https://www.st.com/en/embedded-software/x-cube-ai.html}, accessed:
  2019-09-12.

\bibitem{lai2018cmsis}
L.~Lai, N.~Suda, and V.~Chandra, ``{Cmsis-nn: Efficient neural network kernels
  for arm cortex-m cpus},'' \emph{arXiv preprint arXiv:1801.06601}, 2018.

\bibitem{zhang2017hello}
Y.~Zhang, N.~Suda, L.~Lai, and V.~Chandra, ``Hello edge: Keyword spotting on
  microcontrollers,'' \emph{arXiv preprint arXiv:1711.07128}, 2017.

\bibitem{chowdhery2019visual}
A.~Chowdhery, P.~Warden, J.~Shlens, A.~Howard, and R.~Rhodes, ``Visual wake
  words dataset,'' \emph{arXiv preprint arXiv:1906.05721}, 2019.

\bibitem{garofalo2019pulp}
A.~Garofalo, M.~Rusci, F.~Conti, D.~Rossi, and L.~Benini, ``Pulp-nn:
  Accelerating quantized neural networks on parallel ultra-low-power risc-v
  processors,'' \emph{arXiv preprint arXiv:1908.11263}, 2019.

\bibitem{lecun2015lenet}
Y.~LeCun \emph{et~al.}, ``Lenet-5, convolutional neural networks,'' \emph{URL:
  http://yann. lecun. com/exdb/lenet}, vol.~20, p.~5, 2015.

\bibitem{culurciello}
\BIBentryALTinterwordspacing
``{Continual Learning},'' 2018. [Online]. Available:
  \url{https://medium.com/@culurciello/continual-learning-da7995c24bca}
\BIBentrySTDinterwordspacing

\bibitem{lomonaco2019continual}
V.~Lomonaco, ``Continual learning with deep architectures,'' Ph.D.
  dissertation, alma, 2019.

\bibitem{rusci2018work}
M.~Rusci, A.~Capotondi, F.~Conti, and L.~Benini, ``Work-in-progress: Quantized
  nns as the definitive solution for inference on low-power arm mcus?'' in
  \emph{2018 International Conference on Hardware/Software Codesign and System
  Synthesis (CODES+ ISSS)}.\hskip 1em plus 0.5em minus 0.4em\relax IEEE, 2018,
  pp. 1--2.

\bibitem{rusci2019memory}
M.~Rusci, A.~Capotondi, and L.~Benini, ``Memory-driven mixed low precision
  quantization for enabling deep network inference on microcontrollers,''
  \emph{arXiv preprint arXiv:1905.13082}, 2019.

\bibitem{stm32l4}
\BIBentryALTinterwordspacing
``{STMicroelectronics STM32L476xx},'' 2019. [Online]. Available:
  \url{https://www.st.com/resource/en/datasheet/stm32l476je.pdf}
\BIBentrySTDinterwordspacing

\end{thebibliography}

\end{document}